\documentclass[conference]{IEEEtran}
\IEEEoverridecommandlockouts
\usepackage[draft]{hyperref}
\usepackage[noblocks]{authblk}
\usepackage{marvosym}
\usepackage{subfigure} 
\usepackage{cite}
\usepackage{amsmath,amssymb,amsfonts}
\usepackage{algorithmic}
\usepackage{float}
\usepackage{graphicx}
\usepackage{textcomp}
\usepackage{xcolor}
\usepackage{mathrsfs}
\usepackage{multirow}
\usepackage{tabu}
\usepackage{makecell}
\usepackage{threeparttable}

\def\BibTeX{{\rm B\kern-.05em{\sc i\kern-.025em b}\kern-.08em
    T\kern-.1667em\lower.7ex\hbox{E}\kern-.125emX}}
\begin{document}

\title{ Bone Marrow Cell Recognition: Training Deep Object Detection with A New Loss Function }

\author{Dehao Huang$^1$, Jintao Cheng$^3$, Rui Fan$^{4, 5}$, Zhihao Su$^{1}$, Qiongxiong Ma$^{1,*}$, Jie Li$^{2,*}$, \\
$^{1}$Guangdong Provincial Key Laboratory of Nanophotonic Functional Materials and Devices, \\ 
School of Information and Optoelectronic Science and Engineering, South China Normal University, \\
Guangzhou 510006, China.\\
$^{2}$Department of Hematology, Nanfang Hospital, Southern Medical University, Guangzhou 510515, China.\\
$^{3}$School of Physics and Telecommunication Engineering, South China Normal University, \\
Guangzhou 510006, China.\\
$^{4}$ College of Electronic and Information Engineering, Tongji University, Shanghai 201804, P. R. China. \\
$^{5}$ Shanghai Research Institute for Intelligent Autonomous Systems, Shanghai 201210, P. R. China. \\
Email: huangdehao919@gmail.com, chengjintao@unity-drive.com, rui.fan@ieee.org, suuuzh@sina.com, \\
maqx@m.scnu.edu.cn, ljjean@163.com
}


\maketitle
\let\thefootnote\relax\footnotetext{ This work was supported by Special Funds for the Cultivation of Guangdong College Students' Scientific and Technological Innovation ("Climbing Program" Special Funds) (No. pdjh2021b0134); Key-Area Research and Development Program of Guangdong Province (No. 2020B090922006); Featured Innovation Project of Guangdong Education Department (No. 2019KTSCX034); Science and Technology Program of Guangzhou (No. 202002030165, 2019050001). }
\let\thefootnote\relax\footnotetext{ $ ^* $ Corresponding author. }

\begin{abstract}
For a long time, bone marrow cell morphology examination has been an essential tool for diagnosing blood diseases. However, it is still mainly dependent on the subjective diagnosis of experienced doctors, and there is no objective quantitative standard. Therefore, it is crucial to study a robust bone marrow cell detection algorithm for a quantitative automatic analysis system. Currently, due to the dense distribution of cells in the bone marrow smear and the diverse cell classes, the detection of bone marrow cells is difficult. The existing bone marrow cell detection algorithms are still insufficient for the automatic analysis system of bone marrow smears. This paper proposes a bone marrow cell detection algorithm based on the YOLOv5 network, trained by minimizing a novel loss function. The classification method of bone marrow cell detection tasks is the basis of the proposed novel loss function. Since bone marrow cells are classified according to series and stages, part of the classes in adjacent stages are similar. The proposed novel loss function considers the similarity between bone marrow cell classes, increases the penalty for prediction errors between dissimilar classes, and reduces the penalty for prediction errors between similar classes. The results show that the proposed loss function effectively improves the algorithm's performance, and the proposed bone marrow cell detection algorithm has achieved better performance than other cell detection algorithms.
\end{abstract}

\begin{IEEEkeywords}
Bone marrow smear, Deep learning, Loss function, Cell detection algorithm, YOLOv5.
\end{IEEEkeywords}

\section{Introduction}
\label{sec.introduction}
\IEEEPARstart{T}he morphological examination of bone marrow cells plays a vital role in the diagnosis of blood diseases. It can identify the morphological characteristics of blood diseases such as acute leukemia, hemolymphoid tumors, and dysplasia\cite{MedicalKnowledge2009, MedicalKnowledge2014}. However, the analysis of cell morphology is mainly done manually by experienced hematologists now. This process is not only cumbersome and labor-intensive but also subjective factors that will bring negative effects\cite{Wu2020}. Therefore, a computer-assisted quantitative automatic analysis system is essential, and the cell detection algorithm as its core has aroused the interest of many researchers.

In recent years, the research on cell detection algorithms can be divided into traditional methods based on image processing and shallow machine learning, and methods based on deep learning. The general steps of traditional methods include cell segmentation, feature extraction, and cell classification. In some previous studies, image processing algorithms such as intensity clustering, watershed transformation, and adaptive thresholding were used as segmentation methods\cite{Deep10class}, traditional classifying algorithms such as Support Vector Machine (SVM) and Naive Bayes Classifier were used as cell classifiers\cite{TraditionBayers, TraditionANN}. Although they have achieved excellent results, they are still not enough to apply automatic analysis of bone marrow smear. Traditional methods have the limitations of low learning ability and the need for artificial features\cite{Deep10class}. Algorithms based on deep learning do not need to manually design features but learn features from a large amount of data and have powerful feature extraction capabilities. At the same time, deep learning methods can also be mixed and manually adjusted to maintain strong feature extraction capabilities while maintaining strong feature extraction capabilities. Presently, in some studies on cell classification and detection algorithms based on deep learning, they have achieved better results than traditional methods\cite{Deep10class, Deep5detect, Deep40class, Deep5class, Deep11detect}. Therefore, this paper focuses on a bone marrow cell detection algorithm based on deep learning.

\begin{figure*}[!t]
	\begin{center}
		\centering
		\includegraphics[width=0.95\textwidth]{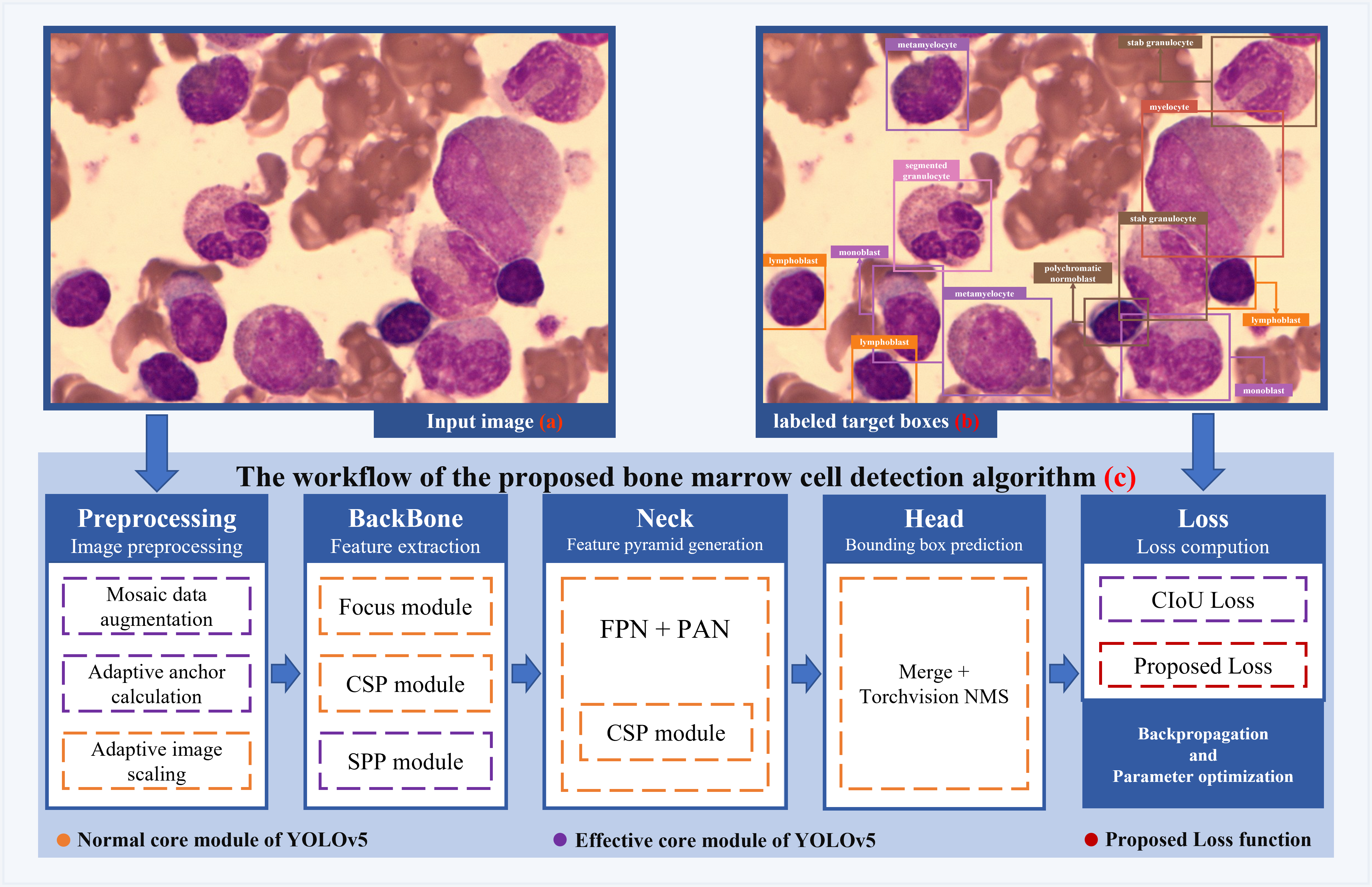}
		\caption{Including the following content: the structure of the proposed bone marrow cell detection algorithm, sub-modules, and the process of using samples to train the model. CSP module means Cross Stage Partial module, SPP module means Spatial Pyramid Pooling module, FPN means module of Feature Pyramid Networks, PAN means module of Pixel Aggregation Network, NMS means Non-maximum suppression, and CIoU means Complete IoU\cite{YOLOv5}. Part (a) is the image of the local area of the bone marrow smear, Part (b) is the sample label corresponding to the image (including cell location, class), and Part (c) is the structure diagram of the proposed method. The sub-modules of the proposed algorithm can be divided into the following three categories: Normal core module of YOLOv5, Effective core module of YOLOv5, and Proposed loss function, corresponding to the three color labels at the bottom of the figure. Both the Normal core module of YOLOv5 and the Effective core module of YOLOv5 are important core modules that support the excellent performance of YOLOv5 networks. Among them, the Effective core module of YOLOv5 has outstanding advantages in detecting bone marrow cells. The following is the algorithm training process: 1. the image input algorithm is forwarded to the Prediction part to obtain the detection result of the algorithm, 2. the loss part inputs the detection result to the loss function and the sample label corresponding to the image, 3. finally backpropagates and optimizes the model parameters.}
		\label{fig.Structure}
	\end{center}
\end{figure*}

Currently, cell detection algorithms based on deep learning develop steadily. Kutlu et al. used Faster-RCNN for end-to-end cell detection in 5 classes and achieved excellent results\cite{Deep5detect}. Wang et al. used YOLOv3 and Single Shot MultiBox Detector(SSD) for end-to-end cell detection in 11 classes and achieved outstanding performance\cite{Deep11detect}.In the existing research, the accuracy of the cell detection algorithm is sufficient for the commercial automatic analysis system of peripheral blood smears. However, the cell detection algorithm is still not effective enough to be used in an automatic bone marrow smear analysis system. (The bone marrow cell detection algorithm in the paper refers to the cell detection algorithm used to detect bone marrow cells.) The cell detection algorithm based on deep learning for bone marrow cell detection has the following three difficulties:

\begin{enumerate}
\item Bone marrow cells classes are divided according to the series and stages of the cells, and the number of classes exceeds 30. The classes of adjacent stages belonging to the same series have a high degree of similarity so that it is difficult to distinguish. In contrast, the classes belonging to different series have apparent differences so that it is easier to distinguish. The above-mentioned inter-class relationship contains information that helps the algorithm classify cells, but it is still difficult for the algorithm to make full use of this information.
\item In bone marrow smears, the cells are densely distributed, and there are many adherent cells. The algorithm should have sufficient cell segmentation capabilities.
\item The classification of part of cell classes needs to be combined with the context information of the surrounding cells in the image for classification judgment. The algorithm should have the ability to extract global context information.
\end{enumerate}

In view of the above three problems, the main contributions of this paper include the following two points:
\begin{enumerate}
\item This paper proposes a novel loss function. The fine-grained classes are defined as finely divided classes according to the principle of different series and different stages. The coarse-grained classes are defined as roughly divided classes according to the principle of merging fine-grained classes of the same series and adjacent stages. A coarse-grained class often contains more than one fine-grained class, and the algorithm directly predicts the fine-grained classes. The proposed loss function will increase the penalty for the model's prediction of coarse-grained classification errors during algorithm training, try to use the similarity between fine-grained classes belonging to the same coarse-grained class to strengthen the supervision of model feature extraction.
\item This paper proposes a bone marrow cell detection algorithm based on YOLOv5 network\cite{YOLOv5}, trained with the novel loss function. The proposed algorithm can segment cells effectively because of powerful feature extraction capabilities, and also can extract context information because of a sub-module that can expand the receptive field. Compared with the existing cell detection algorithms, the proposed bone marrow cell detection algorithm has superior performance.
\end{enumerate}

\section{Related Works}



Cell detection algorithms based on deep learning can be divided into two types. One type of algorithm includes two steps of segmentation and classification: the two-step algorithm. The other type of algorithm completes the end-to-end detection: the end-to-end algorithm. Then the following are analyzed separately.

The two-step cell detection algorithms are relatively common, which first uses the cell segmentation algorithm to segment the image of a single cell from the image and then uses the cell classification algorithm to classify each single cell image. For example, the method in \cite{TraditionDeepMethod1} uses a threshold segmentation algorithm to segment the single-cell image and then uses the Alexnet image classification model to classify the single-cell image. The method in \cite{DeepDeepMethod1} uses Faster-RCNN network to segment the single-cell image and then uses Alexnet to perform fine-grained classification. For cell detection tasks, the segmentation and classification algorithms included in the two-step algorithm are relatively independent and clear to optimize separately. The excellent segmentation algorithm can be combined with the excellent classification algorithm \cite{Deep10class, Deep40class} for cell detection. However, it will have the following limitations when the two-step algorithm is used to detect bone marrow cells: 1. the classification of single cells is challenging to combine context information, 2. the training process is cumbersome, 3. the detection speed is low.

With the rapid improvement of end-to-end object detection network performance, many researchers have begun to use end-to-end cell detection algorithms that are mainly based on end-to-end target detection networks. For example, the method in \cite{Deep5detect} uses the Faster-RCNN network to implement an end-to-end cell detection algorithm and trains them used transfer learning methods. The method in \cite{Deep11detect} is based on YOLOv3 and SSD to achieve end-to-end cell detection algorithms. Compared with the two-step cell detection algorithm, the end-to-end cell detection algorithm has the advantages of a simple training process and fast running speed. What's more, it is relatively easy to combine the contextual information. At present, the main limitation of the end-to-end cell detection algorithm is that it cannot make full use of the characteristics of the bone marrow cell detection task, such as the similarities between the types of bone marrow cells mentioned in Chapter \ref{sec.introduction}.

\section{Methods}
\label{sec. methods}

The YOLOv5 network is a classic one-stage detection network, and it is the latest work of the YOLO architecture\cite{YOLO} series. Compared with the previous work YOLOv4 network, YOLOv5 network has achieved better accuracy and faster inference speed, and nearly 90\% has reduced the volume of model parameters. 

This paper proposes a bone marrow cell detection algorithm based on the YOLOv5 network. Figure \ref{fig.Structure} shows the following information: the structure of the algorithm model, essential sub-modules, and training method. The following content of this chapter introduces two types of essential sub-modules, corresponding to the Effective core modules and Proposed loss function in the Figure \ref{fig.Structure}.

\subsection{Effective Core Module}
The following four sub-modules in the proposed method have four advantages in bone marrow cell detection tasks:
\begin{enumerate}
\item SPP module \cite{SPP}. The algorithm adds the SPP module to the Neck part, which helps the feature map merge local and global features. SPP can extract context-related information of bone marrow cells.
\item Mosaic data enhancement. Lack of data is a major challenge in the bone marrow cell detection task. The algorithm adds Mosaic data enhancement at the input end. It uses four pictures to be spliced by random scaling, random cropping, and random arrangement to enrich the data set.
\item Adaptive anchor boxes calculation. Bone marrow cells are mainly circular, and most of the cells are similar in shape and size. The algorithm adds an adaptive anchor boxes calculation at the input end, making it easy to find the best anchor boxes to improve the model’s ability to locate cells.
\item CIoU loss\cite{CIOU}. In the labeled samples of the bone marrow cell detection task, the rectangular boxes of the labeled cells are often unable to frame the cells due to manual labeling accurately. However, it can be ensured that the center of labeled boxes and the aspect ratio of human-labeled data is approximately the same as ground truth. The proposed algorithm uses CIoU loss, which gains the advantage in considering the overlap area, center point distance, and aspect ratio, making the penalty for the model more effective.
\end{enumerate}

\subsection{Proposed Loss Function}
The YOLOv5 network originally calculates the loss of each labeled ground-truth bounding box using Eq.\ref{eq.origin_loss}, including three parts, namely positioning loss, confidence loss, and classification loss. The way to calculate classification loss is as shown in Eq.\ref{eq.origin_loss_cls}, where $ S^2 $ is the number of grid cells, B is the number of anchor boxes that each grid cell shall predict, $ I_{ij}^{obj\ } $represents whether the labeled ground-truth bounding box of the current calculation loss belongs to the j anchor box of the i grid cell, and classes represent all the categories of the sample, $ \delta $ means the function of Sigmoid,$ \widehat{p_i}\left(c\right )$ is the true probability of class c. In contrast, $ {\ p}_i\left(c\right) $is the predicted probability of class c. The classification loss part calculates the sum of the classification loss of the labeled ground-truth bounding box and all its corresponding anchor boxes.

\begin{equation}
Loss=Loss_{box}+Loss_{obj}+Loss_{cls}
\label{eq.origin_loss}
\end{equation}

\begin{footnotesize} 
\begin{equation}
\begin{aligned}
Loss_{cls}&=\sum_{i=0}^{S\times S}\sum_{j=0}^{B}{I_{ij}^{obj}\times} \\
&\sum_{c\in c l a s s e s}\left[\widehat{p_i}\left(c\right)log{\left(\delta\left(p_i\left(c\right)\right)\right)} + \left(1-\widehat{p_i}\left(c\right)\right)log{\left(1-\delta\left(p_i\left(c\right)\right)\right)}\right]
\end{aligned}
\label{eq.origin_loss_cls}
\end{equation}
\end{footnotesize} 

The experiment found that if we only train the ability of the proposed algorithm to detect the position of bone marrow cells, the algorithm performs exceptionally well, with accuracy and recall reaching over 99\%. Therefore, the limitation of the proposed bone marrow cell detection algorithm lies in the classification of bone marrow cells instead of segmentation.

Based on the above judgment, the weights of each part of the loss function of the YOLOv5 network were adjusted first. A basic weights parameter $ \alpha $ is set to satisfy $ \alpha>1 $ , and the three-part weights of the loss function of the YOLOv5 network is adjusted according to Eq.\ref{eq.optimized_loss} to make it more sensitive to classification loss, which make algorithm focus more on learning classification problems.

\begin{equation}
Loss=Loss_{box}+Loss_{obj}+{\alpha}Loss_{cls}
\label{eq.optimized_loss}
\end{equation}

Next, try to optimize the loss function by using the similarity between the fine-grained classes of bone marrow cells. In the fine-grained classes, the cell classes of the same series at different stages have common characteristics. However, when calculating the classification loss value, the prediction errors of different series of cells and the prediction errors of the same series of cells are equivalent to the penalty effect of the model, which is not conducive to the supervised model to learn more useful features. The proposed loss tries to use the following methods to solve the above problems: Using the proposed loss function Eq.\ref{eq.optimized_loss_cls} to replace the classification part of the loss function Eq.\ref{eq.optimized_loss}. $ \gamma $ in Eq.\ref{eq.optimized_loss_cls} is the same as in Eq.\ref{eq.gammer_value}. When the class of the labeled ground-truth bounding box is inconsistent with the coarse classification class predicted by the j anchor box of i grid cell, the classification loss of the anchor box will be multiplied by the weights $ \left(1+\beta\right) $ to satisfy $ \beta>0 $.
\begin{footnotesize} 
\begin{equation}
\begin{aligned}
{Loss}_{cls}&=\sum_{i=0}^{S\times S}\sum_{j=0}^{B}{I_{ij}^{obj}}\left(1+\gamma\right)\times \\
&\sum_{c\in c l a s s e s}\left[\widehat{p_i}\left(c\right)log{\left(\delta\left(p_i\left(c\right)\right)\right)}+\left(1-\widehat{p_i}\left(c\right)\right)log{\left(1-\delta\left(p_i\left(c\right)\right)\right)}\right]
\end{aligned}
\label{eq.optimized_loss_cls}
\end{equation}
\end{footnotesize}

\begin{equation}
\gamma=\left\{
\begin{array}{rcl}
\beta   & & predict_{coarse}\neq target_{coarse} \\
0       & & predict_{coarse}=    target_{coarse}\\
\end{array} \right.
\label{eq.gammer_value}
\end{equation}

In Eq.3 and Eq.4, $ \alpha $ and $ \beta $ are two hyperparameters. Increasing $ \alpha $ can increase the model's attention to classification, and increasing $ \beta $ can increase the penalty weight for prediction errors of coarse-grained classes.

\section{Experimental Results and Discussion}
\label{sec.experimental_results_and_discussion}
\begin{figure}[H]
	\begin{center}
		\centering
		\includegraphics[width=0.4\textwidth]{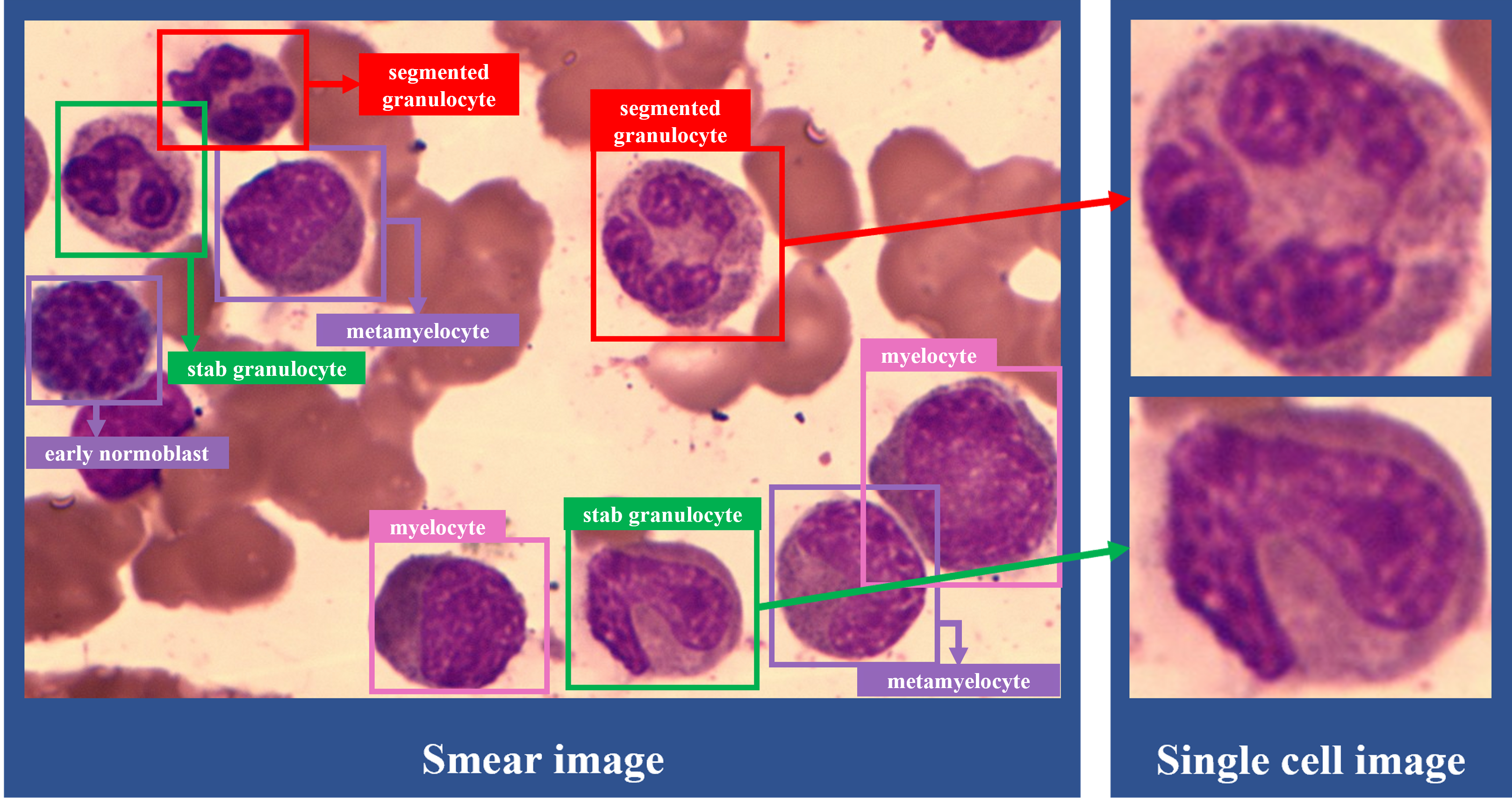}
		\caption{Example of data set}
		\label{fig.dataset}
	\end{center}
\end{figure}

The bone marrow cell labeling data set we used for training and testing came from the Southern Hospital of Southern Medical University and was collected by optical microscope and camera. The data set included 2,549 labeled images from 144 patients, 15,642 labeled individual cells. The size of each image is 1024 x 684, and the total number of types of bone marrow cells is 36. (An example is shown in Figure \ref{fig.dataset}.) To ensure a low correlation between the training set and the test set, we divide the data set with patients as the division unit, the training set contains data of 117 patients, and the test set contains data of 27 patients.

\begin{table*}[!htbp]
\caption{ Experimental results}
\label{table_1}
\renewcommand\arraystretch{1.6}
\centering
	\begin{threeparttable}
		\begin{tabular}{|c|c|c|c|}
		\hline

		\multicolumn{2}{|c|}{ Method and Hyperparameters} & \makecell[c]{mAP@0.5 \\ fine-grained classification} & \makecell[c]{mAP@0.5 \\ coarsed-grained classification} \\
		\cline{1-4}
		\hline

		\multicolumn{2}{|c|}{ SSD\cite{Deep11detect} } & 21.18\% & 37.38\% \\
		\hline

		\multicolumn{2}{|c|}{ Faster-RCNN\cite{Deep5detect} } & 34.97\% & 56.83\% \\
		\hline

		\multicolumn{2}{|c|}{ YOLOv5 + Normal loss function\tnote{1} } & 44.65\% & 66.00\% \\
		\hline

		\multirow{5}*{ \makecell[c]{YOLOv5 + \\ Class weighted loss function\tnote{2}} } & $ \alpha = 2.50 $ & 48.93\% & 66.68\%\\
		\cline{2-4}
		& $ \alpha = 2.75 $ & 48.48\% & 66.77\%\\
		\cline{2-4}
		& $ \alpha = 3.00 $ & 48.76\% & 66.28\%\\
		\cline{2-4}
		& $ \alpha = 3.25 $ & 48.50\% & 67.00\%\\
		\cline{2-4}
		& $ \alpha = 3.50 $ & 49.68\% & 66.14\%\\
		\hline

		{\makecell[c]{YOLOv5 + \\ Proposed loss function\tnote{3}}} & {$\alpha = 2$,  $\beta = 1$} & \textbf{51.14\%} & \textbf{69.71\%} \\
		\hline

		\end{tabular}
		\begin{tablenotes}    
			\footnotesize               
			\item[1] The normal loss is the original loss function of YOLOv5 network.       
			\item[2] The class weighted loss is the loss function that increases the weight of the classification part.        
			\item[3] The proposed loss is the loss function proposed in the paper       
		\end{tablenotes} 
	\end{threeparttable}
\end{table*}









\begin{figure}[H]
	\begin{center}
		\centering
		\includegraphics[width=0.4\textwidth]{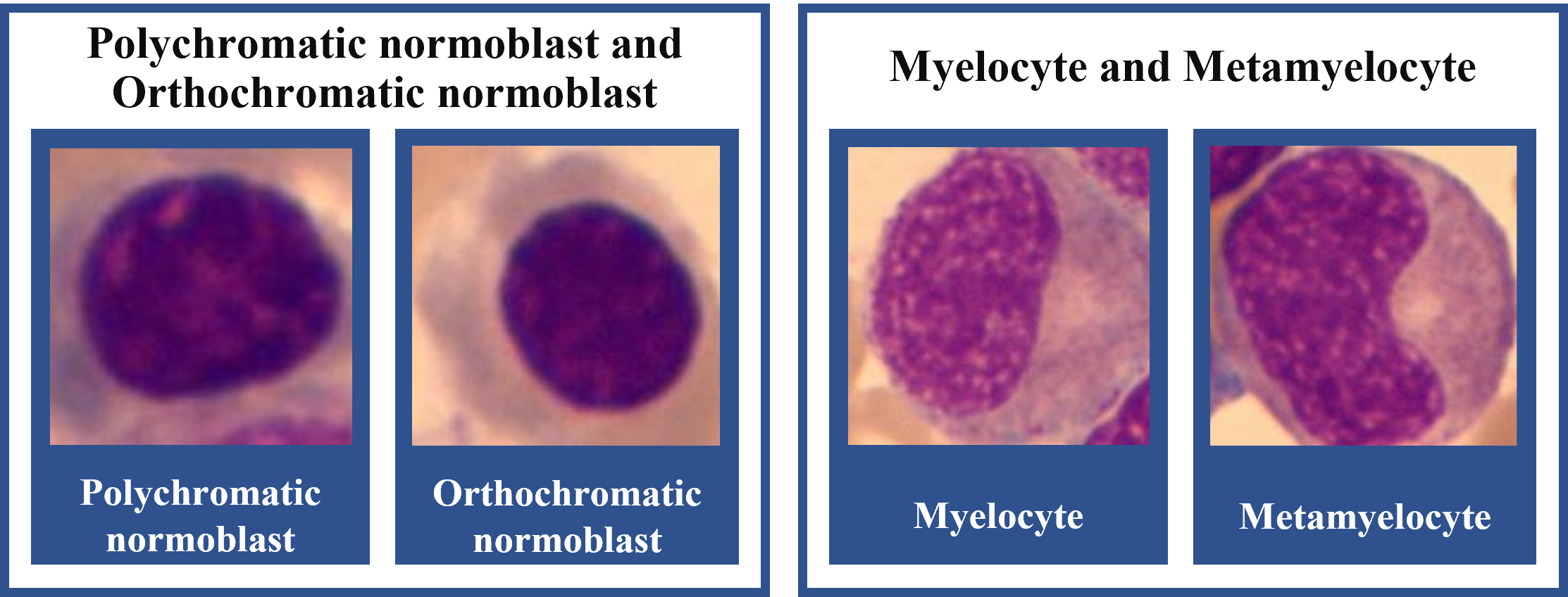}
		\caption{Example of fine-grained classes merged into coarse-grained classes. Polychromatic normoblast and Orthochromatic normoblast belonging to the coarse-grained classes include Polychromatic normoblast and Orthochromatic normoblast belonging to the fine-grained classes, and Myelocyte and Metamyelocyte belonging to the coarse-grained classes include Myelocyte and Metamyelocyte belonging to the fine-grained classes.}
		\label{fig.combine_class}
	\end{center}
\end{figure}

The data set included 36 fine-grained classes, including lymphocyte, Promyelocyte, megaloblasts, Metamyelocyte, megaloblasts, late megaloblasts, etc. Based on the principle of the same series, adjacent stages, the existence of common features, and the need for cell counting when diagnosing blood diseases, we merged 36 fine-grained classes and obtained 14 coarse-grained classes(An example is shown in Figure \ref{fig.combine_class}).

We use the Pytorch deep learning framework and NVIDIA RTX 2070 GPU for training and testing in the following experiments. When experimenting with the methods of other papers, we use the hyperparameters proposed in those papers. When experimenting with the proposed bone marrow cell detection algorithm in this paper, the following hyperparameters are set for training: batch size is 18, total training epochs is 80, and the image size after resize is 640x640.

In order to fully demonstrate the advantages of the algorithm proposed in this paper, we have selected the following two indicators to analyze the performance of the cell detection algorithm quantitatively: 
\begin{enumerate}
\item The algorithm predicts the mAP(mean average precision) indicator of fine-grained classes \cite{mAP}, IOU threshold is set to 0.5, after this referred to as mAP@0.5. 
\item The algorithm predicts the mAP@0.5 indicator of the coarse-grained class. It should be added that both the classes of algorithm prediction and sample annotation are fine-grained classes. The result of the algorithm predicting the coarse-grained classes is obtained from the mapping rule of the fine-grained classes to the coarse-grained classes. 
\end{enumerate}

Using the data set of this paper, the cell detection algorithms which have excellent performance in the current field\cite{Deep5detect, Deep11detect} and the bone marrow cell detection algorithm proposed in this paper are trained, measured, and evaluated. Each experiment was repeated three times to take the average value to avoid the contingency, the indexes obtained in Table \ref{table_1}.

Since the weight of the classification part of the proposed loss function is unevenly distributed between $ \alpha $ and $ \alpha+\beta $, we selected five  $ \alpha $ which are equidistant distributed between $ \alpha $ and $ \alpha(1 + \beta) $ in the class weighted experiment, verifying that proposed loss is not the result of adjusting the weight of the classification section.

From the experimental results in Table \ref{table_1}, the following conclusions can be analyzed:
\begin{enumerate}
\item Proposed bone marrow cell detection algorithm performs well. The mAP@0.5 in the case of fine-grained classification reaches 51.14\%. The mAP@0.5 in the case of coarse-grained classification reaches 69.71\%, which is significantly better than the other two methods\cite{Deep5detect, Deep11detect}. Compared with the Faster-RCNN combined with transfer learning method\cite{Deep5detect}, the proposed algorithm improves the mAP@0.5 of fine-grained classification by 16.17\% and the mAP@0.5 of coarse-grained classification by 12.88\%.
\item Increase the weight of the classification part in the loss function, the model can focus more on the classification of cells, which can effectively improve the performance of the bone marrow cell detection algorithm. Compared with the model with normal loss function, the model with class weighted loss function improves the mAP@0.5 of fine-grained classification by 5.03\%.
\item Use the loss proposed in this paper to adjust the penalty for model prediction of coarse-grained class errors and fine-grained class errors, which can effectively improve the performance of the YOLOv5 network for bone marrow cell detection algorithms. Compared to model with class weighted loss function, the model with proposed loss function improves the mAP@0.5 of fine-grained classification by 1.46\%, and the mAP@0.5 of coarse-grained classification by 3.57\%. The results show that the use of the relationship between the coarse-grained classes and the fine-grained classes to strengthen the supervised training of the model is likely to help the algorithm learn the characteristics of the cell classes more effectively.
\end{enumerate}

It should be noticed that when using cell morphology to diagnose patients, some common blood diseases can determine the diagnosis results by coarse-grained classification and counting of cells. The proposed algorithm effectively improves the model's performance in predicting the coarse classes. , Which improves the feasibility of the bone marrow cell detection algorithm in the automatic analysis system.

In the end, we evaluated the running speed of the proposed bone marrow cell detection algorithm: The model deployed on openvino uses Intel i7-10700 CPU to detect cells in a single image. It takes 73ms, which can reach the requirements of most automatic analysis systems for the efficiency of cell detection algorithms.

\section{Conclusion and Future Work}
\label{sec.conclusion_future_work}
The bone marrow cell detection algorithm is the core of the meaningful bone marrow cell morphology automatic analysis system. This paper proposes a bone marrow cell detection algorithm based on the novel loss function and YOLOv5 network. The experimental results show that the proposed bone marrow cell detection algorithm has a better performance than other cell detection algorithms. The final mAP@0.5 of the fine-grained classification reaches 51.14\%, and the mAP@0.5 of the coarse-grained classification reaches 69.71\%. In future research, we will continuously try to combine model optimization and increase the sample size of the data set to improve the algorithm's accuracy in classifying cells.

\bibliographystyle{IEEEbib}
\bibliography{ref}

\end{document}